\documentclass{article}
\usepackage{float}
\usepackage[colorlinks]{hyperref}
\hypersetup{citecolor=orange}

\usepackage[preprint]{neurips_2019}
\usepackage[utf8]{inputenc} 
\usepackage[T1]{fontenc}    
\usepackage{hyperref}       
\usepackage{url}            
\usepackage{booktabs}       
\usepackage{amsfonts}       
\usepackage{nicefrac}       
\usepackage{microtype}      
\usepackage{amssymb,xcolor,trimclip}

\usepackage[ruled,vlined,linesnumbered,noresetcount]{algorithm2e}
\newcommand{\AlgoResetCount}{\renewcommand{\@ResetCounterIfNeeded}{\setcounter{AlgoLine}{0}}}
\newcommand{\AlgoNoResetCount}{\renewcommand{\@ResetCounterIfNeeded}{}}
\newcounter{AlgoSavedLineCount}

\title{Approximate Bayesian Computation via Population Monte Carlo and Classification}

\date{}
\author{%
    Charlie Rogers-Smith \\
  Aalto University \\
  \texttt{charlierogerssmith@gmail.com} \\
   \And
   Henri Pesonen \\
  Aalto University \\
  \texttt{henri.3.pesonen@aalto.fi} \\
  \And
  Samuel Kaski \\
  Aalto University \\
  \texttt{samuel.kaski@aalto.fi} \\
}

\begin{document}
\maketitle

\begin{abstract}
\vspace{5mm}
Approximate Bayesian computation (ABC) methods can be used to sample from posterior distributions when the likelihood function is unavailable or intractable, as is often the case in biological systems. ABC methods suffer from inefficient particle proposals in high dimensions, and subjectivity in the choice of summary statistics, discrepancy measure, and error tolerance. Sequential Monte Carlo (SMC) methods have been combined with ABC to improve the efficiency of particle proposals, but suffer from subjectivity and require many simulations from the likelihood function. Likelihood-Free Inference by Ratio Estimation (LFIRE) leverages classification to estimate the posterior density directly but does not explore the parameter space efficiently. This work proposes a classification approach that approximates population Monte Carlo (PMC), where model class probabilities from classification are used to update particle weights. This approach, called Classification-PMC, blends adaptive proposals and classification, efficiently producing samples from the posterior without subjectivity. We show through a simulation study that Classification-PMC outperforms two state-of-the-art methods: ratio estimation and SMC ABC when it is computationally difficult to simulate from the likelihood.
  
\end{abstract}
 
\section{Introduction}

\subsection{Being Bayesian}

Bayes' theorem states that, given observed data $Y$ and likelihood function $p(Y|\theta)$, the posterior distribution over parameters $\theta$ is

\begin{equation}
p(\theta|Y) = \frac{p(Y|\theta)p(\theta)}{p(Y)}.
\end{equation}

Intuitively, the posterior distribution sees how likely the observed data are at each possible parameter value $\theta$ using the likelihood function, and weighs this according to our prior beliefs about the plausibility of those parameter values, $p(\theta)$. The marginal likelihood $p(Y)$ on the denominator can be viewed as a normalising constant. The Bayesian approach treats parameters $\theta$ as random variables with an associated distribution, as opposed to the assuming that the parameters are fixed but unknown quantities, as in frequentist statistics. Bayesian inference gives a rigorous treatment of uncertainty and is the natural approach for incorporating prior beliefs into inference. Bayesian inference also sidesteps some problems encountered in frequentist inference, such as the violation of the likelihood principle: that all the evidence learned from a sample that is relevant to parameter inference must be contained in the likelihood function. For example, the p-value from a frequentist significance test may often depend on the design of the experiment (see \cite{berger1988likelihood}), and hence inference is not solely done through the likelihood function. 

We consider in this work the problem of generating samples from the posterior density of the model parameters $\theta\in \mathbb{R}^{d}$, given observed data $Y$ and prior $p(\theta)$. Being able to produce samples from a distribution is generally useful: from samples we are able to calculate summary statistics such as the mean, quantiles, and evaluate high-dimensional integrals such as

\begin{equation}p(Y) = \int_{\Theta} p(Y|\theta)p(\theta)d\theta,  \end{equation}

which is the marginal likelihood from (1). The marginal likelihood has many applications and is particularly important for Bayesian model comparison. However, for a high-dimensional parameter space, evaluating the marginal likelihood analytically is a difficult task, and was the main factor for preferring frequentist approaches. Fortunately, the increasing power and availability of automated computation in the latter half of the 20th century has led to the rise of Bayesian inference for solving practical problems; it is now computationally feasible to produce samples from many distributions, and consequently there is a large, growing body of literature on sampling theory (see \cite{gilks1995markov}, for example).

\subsection{Approximate Bayesian Computation}

\subsubsection{Problem Statement}

There exist many models for which the likelihood function is not available or too costly to compute (see \cite{beaumont2002approximate}, for example). Not being able to compute the likelihood function is problematic for Bayesians and frequentists alike, since the computation of the posterior distribution of parameters depends on the likelihood function, as do frequentist maximum-likelihood methods. The following example from wildlife population ecology illustrates this point.

\subsubsection{The Ricker model}

The Ricker model is an ecological model that describes how the size of an animal population evolves over time \cite{ricker1954stock}. At time $t$ with true population $N_{t}$, the observation $y_{t}$ is assumed to be a stochastic observation of $N_{t}$ with distribution $y_{t}|N_{t}, \phi \sim Poisson(\phi N_{t})$, where $\phi$ is a scaling parameter. The dynamics of the true population are described by a stochastic version of the Ricker map, 

\begin{equation}
N_{t+1} = N_{t}\exp \Big\{ r \Big(1-\frac{N_t}{k}\Big)\Big\} + \sigma e_t
\end{equation}

for $t=1, \cdots , T$, where $k$ is the carrying capacity, $e^{(t)}$ are independent standard Guassian random variables, and $N_0$ is the initial population. The model has three parameters $\theta = (r, \sigma, \phi)$, where $\log r$ is related to the growth rate and $\sigma$ controls the stochasticity. 

Given a realisation of this model $y = (y_{0},y_{1},...,y_{T})$, calculating the likelihood density is computationally infeasible, since marginalisation over latent $n_{t}$ is required. To illustrate this point, consider the likelihood of a time series given $(r, \sigma, \phi)$:

\begin{equation}
p(y_{0},y_{1},...,y_{T}|r, \sigma, \phi) = \prod_{t=1}^{T-1}p(y_t|y_{t-1}, r, \sigma, \phi) = 
\end{equation}
\begin{equation}
\prod_{t=1}^{T-1}\int_{N_{t-1}} \int_{N_{t}} p(N_{t-1}|y_{t-1},\phi) \cdot p(N_t|N_{t-1},r, \sigma) \cdot p(y_{t}|N_{t},\phi) = 
\end{equation}

\begin{equation}
\prod_{t=1}^{T-1}\int_{N_{t-1}} \int_{N_{t}} p(N_{t-1}|y_{t-1},\phi) \cdot  \frac{1}{\sqrt{2 \pi \sigma}} \exp  \Bigg\{ \frac{N_t - N_{t-1}\exp \Big\{ r \Big(1-\frac{N_{t-1}}{k}\Big)\Big\}}{2 \sigma^2}  \Bigg\} \cdot \frac{e^{-\phi N_{t}} (\phi N_{t})^{y_{t}}}{(y_{t})!},
\end{equation}

which is not obviously integrable. Hence while writing the likelihood down is possible, it is computationally infeasible to compute it. However, given a parameter value $(r, \sigma, \phi)$, to generate an artificial observation $y$ given parameter value $\theta = (r, \sigma, \phi)$, all that is required is $T$ generations of standard Gaussian and Poisson variables. Therefore, for the Ricker model it is difficult to analytically compute the likelihood function, but trivially easy to simulate from the data-generating distribution $p(\cdot|\theta)$.

\subsubsection{ABC via Rejection Sampling}

Approximate Bayesian computation (ABC) methods \cite{pritchard1999population} are applied to models when the likelihood is not available or too costly to compute, but the data-generating pdf $p(\cdot | \theta)$ is specified implicitly in terms of a stochastic simulator that generates samples $x$ from $p(\cdot | \theta)$ for all $\theta \in \mathbb{R}^{d}$. Hence ABC techniques are widespread across many disciplines, including population genetics \cite{beaumont2002approximate}, evolution and ecology \cite{beaumont2010approximate}.

ABC methods aim to generate samples from the posterior distribution by finding parameter values that simulate data similar to the observed data. Algorithm 1 describes a vanilla rejection-sampling implementation of ABC. Two questions arise from the goal of ABC: how do we define 'similarity' between simulated data and observed data, and how do we efficiently find parameters that yield data similar to the observed data? 

\begin{algorithm}
\SetAlgoLined
\KwData{Observed $Y$}
\KwResult{$N$ samples from $p(\theta | d(X,Y) \leq \epsilon)$}
\For{ $i = 1$ \KwTo $N $}{\label{forins}
  \Repeat{$d(X_{\theta^{*}},Y) \leq \epsilon$}{
    Generate $\theta^{*} \sim p(\theta)$ \;
    Generate $X_{\theta^{*}} \sim p(\cdot|\theta^{*})$ \;
  }
  $\theta^{(i)} \leftarrow \theta^{*}$ \;
}
\caption{Rejection ABC}
\end{algorithm} 

$\textbf{Defining similarity:}$ suppose we simulate data $X \sim p(\cdot|\theta)$. Then in order to compare the simulated data with the observed data, one may have to compute expert-chosen summary statistics from the data $T_{i}(X)$ for $i = 1, 2,..., n$, for example, the mean and variance. For some data, such as time series, without summary statistics it is not obvious how to compare the data: for example, if the task is to infer the variance parameter for the Ricker model, recording the mean of the differences between consecutive population observations is a more useful summary statistic than the mean of the time series. 

After computing summary statistics, a discrepancy measure $d$ is defined, such as the Euclidean distance, between the two datasets, based on the chosen summary statistics: $d(T(X),T(Y))$, or $d_{T}(X,Y)$, with $T(X) = (T_{1}(X),T_{2}(X),...,T_{n}(X))$. 

In the simplest continuous case, a proposal particle is drawn as a parameter from the prior: $\theta^* \sim p(\theta)$, and data are simulated $x_{\theta^{*}} \sim p(\cdot|\theta^{*})$. If $d_{T}(X_{\theta^{*}},Y) \leq \epsilon$, then $\theta^{*}$ is appended to the set of plausible parameters, from which relevant statistics may be calculated. This process is repeated until $N$ particles have been accepted. $\epsilon$ is the tolerance threshold, which is typically tuned to trade off between the number of accepted particles and the quality of their posterior approximation: an accepted particle $\theta^{*}$ is distributed as $p(\theta | d_{T}(X_\theta,Y) \leq \epsilon)$, so for small $\epsilon$ the accepted particle $\theta^{*}$ approximates the true posterior distribution, but many particles are proposed and rejected before one is accepted. For a larger value of $\epsilon$, fewer particles are rejected, but each particle is less similar to the posterior distribution. Hence there is a tradeoff between the portion of accepted particles and the accuracy of their approximation, which is problem-dependent. When choosing $\epsilon$ one may consider the costs of simulation: time taken to simulate from the prior and the likelihood function, along with the benefits of decreasing $\epsilon$: how 'different' are the accepted particles for two different $\epsilon$ choices?

Note that there is subjectivity in the choice of summary statistics, the discrepancy measure and the error threshold, and different choices can lead to greatly differing inferences, as well as requiring much of the researcher's time to experiment and tune.

$\textbf{Efficient proposals:}$ simulating from the prior as above typically results in a high rejection rate, particularly in high dimensions, and thus is too computationally intensive. One approach to finding parameter values that yield data similar to the observed data is to carefully choose the proposal distribution \cite{filippi2013optimality}, so that proposals are believed to be more plausible than random samples from the prior.

\subsection{Motivation and Contributions}

The two main problems with rejection ABC are the subjectivity in the choice of summary statistics, discrepancy measure, and error threshold, as well as the inefficient proposal of particles, which are sampled from the prior. These problems have been solved individually: Likelihood-Free Inference by Ratio Estimation (LFIRE)\cite{dutta2016likelihood} estimates the posterior density directly by classifying between the likelihood function $p(X|\theta)$ and the marginal likelihood $p(X)$. The ratio $p(X|\theta)/p(X)$ is estimated through classification, and by multiplying through by the prior $p(\theta)$ the posterior density is recovered. By considering the model-class probabilities as opposed to a rejection sampling scheme, LFIRE is able to sidestep the subjectivity problem. Sequential Monte Carlo (SMC) ABC \cite{toni2009approximate} adaptively selects particles based on previously accepted particles. Intuitively, it makes more sense to search in the surrounding area of an accepted particle than blindly sampling from the prior. 

However, LFIRE cannot efficiently explore the parameter space, and SMC ABC suffers from the same subjectivity problems visited in rejection ABC. The core contribution of this work is to blend classification and adaptive proposals to solve both problems simultaneously. In the latter part of this paper, we review the two state-of-the-art approaches detailed above, propose a novel approximation to population Monte Carlo (PMC), called Classification-PMP, and compare its performance with the other methods. We find that Classification-PMC outperforms LFIRE while retaining many of its desirable properties, and outperforms SMC ABC, particularly when simulating from $p(\cdot|\theta)$ is intensive. The intended reader of this thesis is a final year BSc Statistics student who is familiar with statistical inference, Bayesian inference, and has an elementary understanding of classification methodology.

\section{Review of Previous Methods}

\subsection{Primer: Importance Sampling}

Much of this work uses ideas from importance sampling, so what follows is a brief review. Suppose there is a function $f(\theta)$ from which we wish to compute $\mu = \mathbb{E}(f(\Theta))$. At a glance it may appear that usual Monte Carlo methods will suffice. Here is a potential approach: sample $\theta_i$ from the prior $p(\cdot)$ for $i = 1,2,...,N$, then compute
\begin{equation}
\mu \approx \frac{1}{N}\sum_{i=1}^{N}f(\theta_i).
\end{equation}

However, if $f(\theta)$ is close to zero outside a region $G \subset \Theta$ for which $p(\theta \in G)$ is small, then it is possibe than none of the $N$ samples are in $G$, and hence the Monte Carlo estimate performs poorly. It may also be difficult to sample from $p(\cdot)$. This motivates importance sampling: instead of sampling from $p(\cdot)$, an important area of $\Theta$ is sampled from using a proposal distribution $q(\theta)$, which has the same support as $f(\theta)$. Then one must correct for the fact that $q(\theta)$ was sampled from, not $f(\theta)$. The precise correction is informed by the following result:

\begin{equation}
\mu = \int_\Theta f(\theta)p(\theta)d\theta = \int_\Theta \frac{f(\theta)p(\theta)}{q(\theta)}q(\theta)d\theta = \mathbb{E}_q\big(\frac{f(\theta)p(\theta)}{q(\theta)}\big),
\end{equation}

where $\mathbb{E}_q(\cdot)$ is the expectation for $\Theta \sim q(\cdot)$. Hence, if we sample $\theta_i \sim q(\cdot)$ for $i = 1,2,...,N$, instead of averaging the $f(\theta_i)$ we compute

\begin{equation}
\mu \approx \frac{1}{N}\sum_{i=1}^{N}\frac{f(\theta_i)p(\theta_i)}{q(\theta_i)} = \frac{1}{N}\sum_{i=1}^{N} w(\theta_i)f(\theta_i),
\end{equation}

where $w(\theta_i) = p(\theta_i)/q(\theta_i)$. Here we weight each sample according to its importance. It is difficult in our problem setting to construct a priori the proposal distribution $q(\theta)$, since the posterior distribution is not known. However, it is possible to leverage previously accepted particles as a basis for constructing the proposal distribution. This is the approach of sequential Monte Carlo ABC.

\subsection{Sequential Monte Carlo ABC}

\begin{algorithm}
\SetAlgoLined
\KwData{Observed $Y$}
\KwResult{$N$ samples from $p(\theta | d(X,Y) \leq \epsilon_{T})$}
$\textbf{Initialization}$
Set a schedule of decreasing tolerance thresholds $\epsilon_{1} \geq \cdots \geq \epsilon_{T}$

\For{ $j = 1$ \KwTo $N $}{\label{forins}
  \Repeat{$d(X_{\theta^{*}},Y) \leq \epsilon_{1}$}{
    Generate $\theta^{*} \sim p(\theta)$ \;
    Generate $X_{\theta^{*}} \sim p(\cdot|\theta^{*})$ \;
  }
  $\theta_{1}^{(j)} \leftarrow \theta^{*}$ \;
  $w_{1}^{(j)} \leftarrow 1/N$ \;
}
\For{ $i = 2$ \KwTo $T $}{\label{forins}
  \For{ $j = 1$ \KwTo $N $}{\label{forins}
    \Repeat{$d(X_{\theta^{*}},Y) \leq \epsilon_{i}$}{
      Sample $\theta^{**}$ from $\theta_{i-1}$ with weights $w_{i-1}$ \;
        Generate $\theta^{*} \sim q(\cdot|\theta^{**})$ \;
        Generate $X_{\theta^{*}} \sim p(\cdot|\theta^{*})$ \;
    }
    $\theta_{i}^{(j)} \leftarrow \theta^{*}$ \; 
    $w_{i}^{(j)} \leftarrow p(\theta^{*})/(\sum_{k=1}^{N} w_{i-1}^{(k)}q(\theta^{*}|\theta_{i-1}^{(k)}) $ \;
  }
}
\caption{Sequential Monte Carlo ABC}
\end{algorithm} 

Adaptive particle proposals are desirable to efficiently explore a high-dimensional parameter space, in which proposing from the prior distribution will result in a high rejection rate. Sequential Monte Carlo ABC (SMC ABC), described in Algorithm 2, uses information from the accepted proposals (i.e. plausible areas of the parameter space) to sample the proposals for the next iteration: instead of proposing particles from the prior, we propose particles based on previous accepted particles. Leveraging importance sampling arguments: if the proposal distribution is $q(\theta)$, then excecuting a rejection sampling scheme will produce samples $\theta \sim q(\theta)Pr(d_{T}(X_\theta,Y) \leq \epsilon)$. By applying weight $w^{(i)}$ to accepted particle $\theta^{(i)}$ as follows,

\begin{equation}
w^{(i)} \propto \frac{p(\theta^{(i)})}{q(\theta^{(i)})},
\end{equation}

the weighted samples are distributed as $p(\theta|d_{T}(X_\theta,Y) \leq \epsilon)$. 

SMC ABC proceeds iteratively: at iteration $i$ generating $N$ samples from $p(\theta | d(X_\theta,Y) \leq \epsilon_{i})$, with decreasing $\epsilon$ schedule: $\epsilon_{1} \geq \cdots \geq \epsilon_{T}$. In the final iteration SMC ABC produces $N$ samples from the approximate posterior distribution $p(\theta | d(X_\theta,Y) \leq \epsilon_{T})$. Because of the decreasing epsilon schedule, at each subsequent iteration the accepted parameter values better approximate the posterior distribution. For particle $\theta^{(i)}$ at iteration $t$, the proposal distribution $q_t(\theta)$ is based on a weighted sum of the particles from the previous iteration

\begin{equation}
q_t(\theta) = \frac{1}{N} \sum_{i=1}^N w_{t-1}^{(i)} q_t(\theta|\theta_{t-1}^{(i)}),
\end{equation}

where $q_t(\theta|\theta_{t-1}^{(i)})$ is typically a multi-dimensional Gaussian distribution with mean $\theta_{t-1}^{(i)}$ and covariance matrix $2Cov(\theta_{t-1})$. More on this later.

Core idea: SMC ABC uses particles accepted at the previous iteration to propose new particles. As the error threshold decreases, the particles more closely resemble the posterior distribution. By proposing particles adaptively, far fewer simulations from the likelihood function are required than if one were to use a vanilla rejection sampling scheme as in section 2.2.3.
 
However, SMC ABC is not without problems: for small values of $\epsilon_{T}$ it is possible for parameter values sampled from the true posterior to be rejected. This is exaggerated when the observed data are not a representative sample of the true distribution, as is common with small datasets. A poor choice of schedule and $\epsilon_{T}$ may result in a high rejection rate of particles. When the data-generating distribution $p(\cdot|\theta)$ is hard to simulate from, generating a large number of samples from the likelihood of each particle at each iteration is infeasible. SMC ABC also suffers from the subjectivity problem: we are still required to choose the summary statistics, discrepancy measure, and an epsilon schedule.

\subsection{Ratio Estimation}

Where ABC produces samples from the posterior distribution, ratio estimation aims to estimate the posterior density directly: the posterior density at a parameter value $\theta$ is approximated by estimating the ratio between the data-generation distribution $p(\cdot|\theta)$ and the marginal distribution $p(x)$ via classification, termed as Likelihood-Free Inference by Ratio Estimation (LFIRE)\cite{dutta2016likelihood}. LFIRE is described by Algorithm 3. By classifying in order to compute the posterior density, LFIRE removes the need to define a discrepancy measure and error threshold, and enjoys automatic selection of relevant summary statistics when the $L_1$ penalty is applied to the loss function - more on this later. Hence LFIRE completely solves the subjectivity problem.

\begin{algorithm}
\SetAlgoLined
\KwData{Observed data $Y$ (such as a set of time series), a set of particles $\theta_i$ for $i = 1, 2,...n$ from which to evaluate the posterior density}
\KwResult{A set of estimates of $p(\theta_i | Y)$ for $i = 1, 2,...n$}
Create the dataset for the marginal likelihood $p(\cdot)$:
\For{ $j = 1$ \KwTo $M$}{\label{forins}
  Generate $\theta^{*} \sim p(\theta)$ \;
  Generate $z_j \sim p(\cdot|\theta^{*})$ \;
}
\For{ $i = 1$ \KwTo $n$}{\label{forins}
  Create the dataset for the likelihood $p(\cdot|\theta_i)$:
  \For{ $j = 1$ \KwTo $N$}{\label{forins}
    Generate $x_j \sim p(\cdot|\theta_i)$ \;
  }
  Train classifier $C(X, Z)$ (where $Z = (z_1,...,z_M)$)\;
  Predict $p(\theta_{t}^{(i)}|Y)$ using $C(X, Z)$ as in (12)
}
\caption{Likelihood-free Inference by Ratio Estimation}
\end{algorithm}

The posterior density at parameter value $\theta$ can be evaluated by estimating the ratio between $p(Y|\theta)$ and $p(Y)$, since the posterior distribution can be written as

\begin{equation}p(\theta|Y) = \frac{p(Y|\theta)}{p(Y)} \cdot p(\theta) = r(Y,\theta) \cdot p(\theta),\end{equation} 

and because the denominator does not depend on $\theta$,

\begin{equation}\hat{L}(\theta) \propto \hat{r}(Y,\theta).  \end{equation}   

Likelihood-free inference can therefore be performed by ratio estimation. While there are many methods for density ratio estimation \cite{sugiyama2012density}, such as density difference, we focus on probabilistic classification between the two distributions. The idea is to simulate a dataset $X_{\theta} \sim p(\cdot|\theta)$ and dataset $X \sim p(x)$, and build a classifier $C_{\theta}(\cdot)$ based on these datasets. In practice, to simulate $X \sim p(\cdot)$, first $\theta^{*}\sim p(\theta)$ is simulated, and then $x^{*} \sim p(\cdot|\theta^{*})$. Once $C_{\theta}$ has been built, the density ratio is recovered via Bayes theorem, where $M^{(0)}: Y \sim p(\cdot)$, $M^{(i)}: Y \sim p(\cdot|\theta^{(i)})$, and $\pi = p(M_0)$

\begin{equation}\frac{p(Y|\theta^{(i)})}{p(Y)} = \frac{p(Y|M^{(1)})}{p(Y|M^{(0)})} = \frac{p(M^{(1)}|Y) p(Y)}{p(M^{(1)})} \Biggm/ \frac{p(M^{(0)}|Y)p(Y)}{p(M^{(0)})} = \frac{p(M^{(1)}|Y)}{p(M^{(0)}|Y)} \cdot \frac{1-\pi}{\pi}, \end{equation}

which factors the ratio in terms of the model-class probabilities. Ratio estimation by classification has many appealing properties; one of which is being able to simulate the datasets and fit the classification model before seeing the observed data, which is only used for the final prediction. This offline feature is useful when waiting for medical data, for example, because predictions should be made as quickly as possible after the data are obtained. Standard ABC approaches simulate data of a similar size to the observed data. This fails when there are few data points. In the LFIRE approach, the accuracy of the classification is dependent only on the number of simulations used for $X$ and $X_{\theta}$. Since $X$ is independent of parameter values it can be reused for each classification, and consequently can be made large in order to improve classification accuracy.

However, LFIRE does not explore the parameter space efficiently: it simulates particles from the prior, and uses the same computational resources at each particle when computing its posterior density. So while LFIRE solves the subjectivity problem, it wastes information gained from estimating the posterior density.

\section{Methods}

\subsection{Population Monte Carlo}

Population Monte Carlo (PMC) methods \cite{cappe2004population} are used to generate samples from the posterior distribution when the posterior density is available. PMC cannot be directly used in conjunction with ABC because ABC is applied in the situations where the likelihood, and therefore the posterior, is not available in analytical form required for the desired weight-update. Hence, in order to apply PMC, we require methods of approximating the particle weights in PMC, which will be presented in the next section. See Algorithm 4 for full details of PMC.

Population Monte Carlo generates, at each iteration, an approximate sample from a target distribution and asymptotically unbiased estimates of integrals under that distribution. PMC constructs the appropriate estimators from particles using importance sampling arguments, constructs the proposal distribution in an MCMC-like framework, and uses ideas from SIR \cite{rubin1987calculation} and iterated particle systems \cite{doucet2001introduction} for sample equalization and improvement.

Samples at each iteration of PMC are constructed using sample-dependent proposals for particle generation as in SMC ABC, but PMC uses importance sampling weights for pruning the sample as opposed to accepting or rejecting a given particle. Another important difference is that while SMC ABC uses a schedule of target distributions, PMC has a static target distribution and produces approximate samples from the target distribution at every iteration: if the sample $\theta_{t}$ is produced by simulating the $\{\theta^{(i)}_t\}$ from distributions $\{q^{(i)}_t\}_{i = 1}^n$, independent of one another and conditional on the past samples, then giving $\theta_{t}^{(i)}$ the importance weight 

\begin{equation}w_{t}^{(i)} = \pi(\theta_{t})/q_{t}^{(i)}(\theta_{t})\end{equation}

results in estimators

\begin{equation}\mathcal{J}_{t} = \frac{1}{N} \sum_{i=1}^{N} w_{t}^{(i)} h(\theta_{t}^{(i)}) \end{equation}

being unbiased for every function $h$ and at every iteration $i$. This is the key property of PMC: it extends importance sampling to proposal distributions for $\theta_{t}^{(i)}$ that can be dependent on both the sample ${i}$ and iteration ${t}$. Therefore all of the samples can be exploited for estimation and adaptation, meaning that the effective sample size is $N \times T$ and consequentially that $N$ need not be large. In reality, PMC requires a few iterations to produce particles that resemble the posterior distribution, since the weights of a small number of particles constitute the majority of the total weight. Finally, the subjectivity observed in SMC ABC is avoided. 

\subsubsection{Practicalities}

We choose $q_{t}^{(i)} = \mathcal{N}(\cdot|\theta_{t-1}^{(i)},\tau^{2})$, with $\tau^{2}$ as twice the weighted empirical covariance of $\theta_{t-1}$. This choice of adaptive proposal distribution was shown by \cite{beaumont2009adaptive} to minimise the Kullack-Leibler (K-L) divergence \cite{kullback1951information} between the target and the proposal distribution. The K-L divergence is a measure of how different one probability distribution $p(\theta)$ is to another probability distribution $q(\theta)$, where $\mathcal{D}_{KL}(p||q) = 0$ implies closeness. It is given by

\begin{equation}\mathcal{D}_{KL}(p||q) = \int_{\Theta} p(\theta)log \Bigg(\frac{p(\theta)}{q(\theta)} \Bigg).  \end{equation}

\begin{algorithm}
\SetAlgoLined
\KwData{Observed $Y$}
\KwResult{$N \cdot T$ samples from $p(\theta|Y)$}
\For{ $i = 1$ \KwTo $N $}{\label{forins}
Generate $\theta_{1}^{(i)} \sim p(\theta)$ \;
Set $w_{1}^{(i)} \leftarrow 1/N$ \;
}
Set $\tau^{2} \leftarrow 2\mathrm{Cov}{\theta_{1}}$ \;
\For{ $t = 2$ \KwTo $T $}{\label{forins}
\For{ $i = 1$ \KwTo $N $}{\label{forins}
Sample $\theta^{*}$ from $\theta_{t-1}$ with weights $w_{t-1}$ \;
Sample $\theta_{t}^{(i)} \sim \mathcal{N}(\theta^{*},\tau^{2})$ \;
Evaluate $p(\theta_{t}^{(i)}|Y)$ \;
$w_{t}^{(i)}\!\leftarrow\!p(\theta_{t}^{(i)}|Y)/(\sum_{k=1}^{N}\!w_{t-1}^{(k)}\mathcal{N}(\theta_{t}^{(i)}|\theta_{t-1}^{(k)},\tau^{2})$ \;
}
Set $\tau^{2} \leftarrow 2\mathrm{Cov}{\theta_{t},w_t}$ (weighted empirical covariance) \;
}
\caption{Population Monte Carlo}
\end{algorithm}

\subsection{Classification-PMC}

Before detailing how to approximate PMC with classification, let us return briefly to the ratio estimation framework. Consider the case where posterior density of $\theta^{(1)}$, $p(\theta^{(1)}|Y)$ has been estimated using LFIRE. In order to make use of all of the generated data from the classification between $p(Y)$ and $p(Y|\theta^{(1)})$, the posterior for $\theta^{(2)}$ may be written as

\begin{equation}p(\theta^{(2)}|Y) =  \frac{p(Y|\theta^{(2)})p(\theta^{(2)})}{p(Y)} = p(Y|\theta^{(2)})p(\theta^{(2)}) \Biggm/ \frac{p(Y|\theta^{(1)})p(\theta^{(1)})}{p(\theta^{(1)}|Y)} \end{equation}
\begin{equation}= p(\theta^{(1)}|Y) \cdot \frac{p(\theta^{(2)})}{p(\theta^{(1)})} \cdot \frac{p(Y|\theta^{(2)})}{p(Y|\theta^{(1)})},
\end{equation}

where the last term may be estimated again through ratio estimation using the datasets generated for estimation of the posterior density for both $\theta^{(1)}$ and $\theta^{(2)}$, meaning that the only computational cost is in fitting the classification model. Each estimation of the likelihood ratio produces two additional posterior density estimates, since $p(\theta^{(1)}|Y)$ can be computed in a similar way. This is extremely useful for cases where it is computationally intensive to simulate from the likelihood function, but comparatively easy to perform classification.

Recall that PMC assigns each particle a weight that is related to the relative plausibility of that particle given the observed data. The likelihood ratio via ratio estimation approach detailed above can be interpreted as an attempt to gain information of the relative plausibility of $\theta^{(1)}$ vs $\theta^{(2)}$. By considering the normalised weight of particle $\theta^{(i)}$, the posterior density calculation can be generalised to multiple particles, and hence omitted:

\begin{equation}
\scalebox{1.5}{$w_i = \frac{\frac{p(\theta^{(i)}|Y)}{q(\theta^{(i)})}}{\sum_{j=1}^{N} \frac{p(\theta^{(i)}|Y)}{q(\theta^{(j)})}} = \frac{\frac{p(Y|\theta^{(i)})p(\theta^{(i)})}{p(Y)q(\theta^{(i)})}}{\sum_{j=1}^{N} \frac{p(Y|\theta^{(j)})p(\theta^{(j)})}{p(Y)q(\theta^{(j)})}} = $}\end{equation} 
\begin{equation}
\scalebox{1.5}{$\frac{p(Y|\theta^{(i)})p(\theta^{(i)})}{q(\theta^{(i)}) \sum_{j=1}^{N} \frac{p(Y|\theta^{(j)})p(\theta^{(j)})}{q(\theta^{(j)})}} = 
\frac{p(\theta^{(i)})}{q(\theta^{(i)}) \sum_{j=1}^{N} \frac{p(\theta^{(j)})}{q(\theta^{(j)})}\frac{p(Y|\theta^{(j)})}{p(Y|\theta^{(i)})}},$}
\end{equation}

which allows for pairwise ratio estimation in the last term of the denominator, leading to $N(N-1)/2$ comparisons across the $N$ parameter values. Alternatively, the unnormalised weight can be formulated in a multi-class classification framework as follows. Let $p(M^{(i)}|Y) = Pr(Y \sim p(\cdot|\theta^{(i)})$. Then 

\begin{equation}w_i = \frac{p(\theta^{(i)}|Y)}{q(\theta^{(i)})} = \frac{p(Y|\theta^{(i)})p(\theta^{(i)})}{q(\theta^{(i)})p(Y)} =  \frac{p(Y|M^{(i)})p(\theta^{(i)})}{q(\theta^{(i)})p(Y)} = \end{equation}
\begin{equation} \frac{p(M^{(i)}|Y)p(Y)p(\theta^{(i)})}{q(\theta^{(i)})p(Y)p(M^{(i)})} = \frac{p(M^{(i)}|Y)p(\theta^{(i)})}{q(\theta^{(i)})p(M^{(i)})}. \end{equation}

This approach allows the weights to be computed when the posterior density is not directly available, as in ABC. The model-class probabilities approximate how likely it is that data $Y$ is distributed as $p(\cdot|\theta^{(i)})$ relative to the other particles. This is the same as the posterior weight given by exact PMC, where the only approximation is the quality of the classification. Using model class-probabilities for the weights within PMC circumvents subjectivity as in LFIRE: since classification is used, Classification-PMC also enjoys automatic selection of summary statistics under the $L_1$ norm. By blending adaptive proposals and classification the parameter space is explored efficiently, since the power of classification is leveraged to inform which particles should be sampled at the next iteration. Algorithm 5 gives more details on Classification-PMC, which is the method that we focus on.

\begin{algorithm}
\SetAlgoLined
\KwData{Observed $Y$}
\KwResult{$N \cdot T$ samples from $p(\theta|Y)$}\
\For{ $i = 1$ \KwTo $N $}{\label{forins}
  Generate $\theta_{1}^{(i)} \sim p(\theta)$ \;
  Set $w_{1}^{(i)} \leftarrow 1/N$ \;
}
Set $\tau^{2} \leftarrow 2\mathrm{Cov}{\theta_{1}}$ \;
\For{ $t = 2$ \KwTo $T $}{\label{forins}
  \For{ $i = 1$ \KwTo $N $}{\label{forins}
    Sample $\theta^{*}$ from $\theta_{t-1}$ with  weights $w_{t-1}$ \;
    Sample $\theta_{t}^{(i)} \sim \mathcal{N}(\theta^{*},\tau^{2})$ \;
  \For{ $j = 1$ \KwTo $M $}{\label{forins}
    Simulate $X_{i,j} \sim p(\cdot|\theta_{t}^{(i)})$  \;
  }
}
Train multi-class classifier $C(X)$

\For{ $i = 1$ \KwTo $N $}{\label{forins}

Calculate $w_{t}^{(i)}$ from $C(X)$ as in equations 22-23, with 

$q(\theta^{(i)}) = \sum_{k=1}^{N} w_{t-1}^{(k)}\mathcal{N}(\theta_{t}^{(i)}|\theta_{t-1}^{(k)},\tau^{2})$
}
Set $w_{t} \leftarrow w_{t}/\sum w_{t}$

Set $\tau^{2} \leftarrow 2 Cov(\theta_{t},w_t)$ (weighted empirical covariance) 
}
\caption{Classification-PMC}
\end{algorithm}

\pagebreak

\section{Experiments}

\subsection{Model: Multivariate Gaussian}

The comparison of Classification-PMC to LFIRE and SMC ABC is illustrated on a multivariate Gaussian inference problem: a 5-dimensional multivariate Gaussian with mean $\mu = (\mu_{1}, \mu_{2}, \mu_{3}, \mu_{4}, \mu_{5}) = (1,2,3,4,5)$ and covariance matrix equal to the identity matrix. A uniform prior $\mathcal{U}(-10, 10)$ is assumed for each $\mu_{i}$. The observed data $Y$ is a single sample from the distribution, to make learning hard, and the sample means are used as summary statistics. In this somewhat unique case, it is not necessary to compute the summary statistics, since they are just the observation. The likelihood of this model is analytically available, so while the methods described in this thesis are not necessary for inference, having access to the likelihood function makes a comparison to the exact weights in PMC possible. R 1.2.1114 was used for the construction of LFIRE, PMC, and Classification-PMC, as well as to produce plots for the results. Python 3.2.8 was used to obtain the results for SMC ABC. Code may be found in the Appendix.

\subsection{Comparison to LFIRE}

LFIRE produces posterior density estimates for a set of parameters. From the density estimates it is possible to approximate PMC, where the estimate given by ratio estimation is used in place of the exact posterior density. Therefore a PMC framework is used to compare LFIRE and Classification-PMC, since it is possible to control for the number of particles, the number of simulations per particle, and the number of iterations.

\subsubsection{Classification Methods}

\textbf{Penalised Logistic Regression:} Ratio estimation may be used in conjunction with any classifier. However, non-linear logistic regression \cite{friedman2001elements} has been shown to work well on problems similar to ours \cite{gutmann2012noise}, and it is possible to achieve automatic selection of summary statistics if the classification is done via non-linear logistic regression: recall that to perform the classification, summary statistics are often computed from the data, which can be subjective. Using the LASSO penalty on the coefficients of summary statistics in the parameterising function \cite{tibshirani1996regression} shrinks the coefficients of unimportant summary statistics to $0$, producing a sparse set of relevant statistics and thus removing the subjectivity in the choice of summary statistics. A full discussion of the LASSO penalty is beyond the scope of this work. 

\textbf{Multinomial Logistic Regression:} Multinomial logistic regression is the generalisation of logistic regression to multiple classes. While any multi-class classification algorithm can be used for Classification-PMC, the most direct comparison to the LFIRE approach was chosen, i.e. the multi-class generalisation. The relevant summary statistics were already given for our experimental model, and hence the application of the $L_{1}$ norm was not necessary.

\subsubsection{Results}

We first compare the performance of the weights assigned by LFIRE and Classification-PMC for the 5-dimensional multivariate Gaussian model. The K-L divergence between the approximate weights and the true weights is monitored. Comparing to the true distribution is possible because we have access to the likelihood function and hence the exact weights in PMC for the multivariate Gaussian model. The K-L divergence is computed over 100 datasets, each consisting of a single observation randomly sampled from $\mathcal{U}(-10, 10)$ for each $\mu_{i}$, and performed classification with 100 simulations from the likelihood to construct the datasets for each particle. In Figure 1 the number of particles to which the weights are assigned is varied. 

\begin{figure}
    \centering
    \caption{Boxplots of the resultant K-L divergences between the approximate weights and exact weights when Classification-PMC and LFIRE are applied across 100 datasets for different numbers of particles. We conclude that Classification-PMC is more accurate than LFIRE for all particle configurations.}
    \includegraphics[trim=0cm 0.2cm 1.8cm 0cm, width=0.45\textwidth]{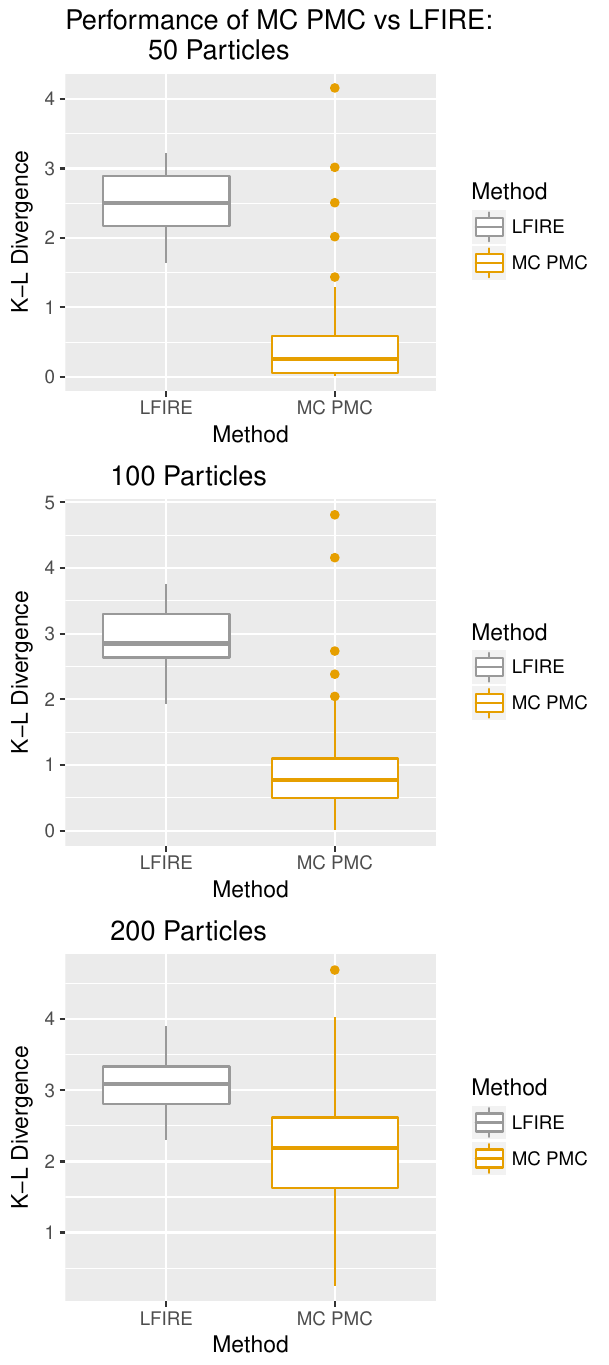}
\end{figure}

It is clear that Clasification-PMC is a closer approximation to the true weights for all particle configurations. A paired t-test rejects the hypothesis that the true difference in means is equal to zero with p-value less than $10^{-11}$, which is strong evidence for the suitability of multi-class classification over LFIRE within PMC. However, Figure 2 shows that Classification-PMC is quadratic in the number of particles for the time taken to perform the classification, while LFIRE is linear in the number of particles. This is because there are $O(N^2)$ comparisons between $N$ particles in multi-class classification.

\begin{figure}[p]
    \centering
    \caption{Boxplots of the time required to fit the multi-class classification model when Classification-PMC is applied across 100 datasets for varied particle sizes. As expected, there is a quadratic relationship between the number of particles and the time taken to train Classification-PMC.}
    \includegraphics[trim=0.4cm 0cm 0cm 0cm, width=1\linewidth]{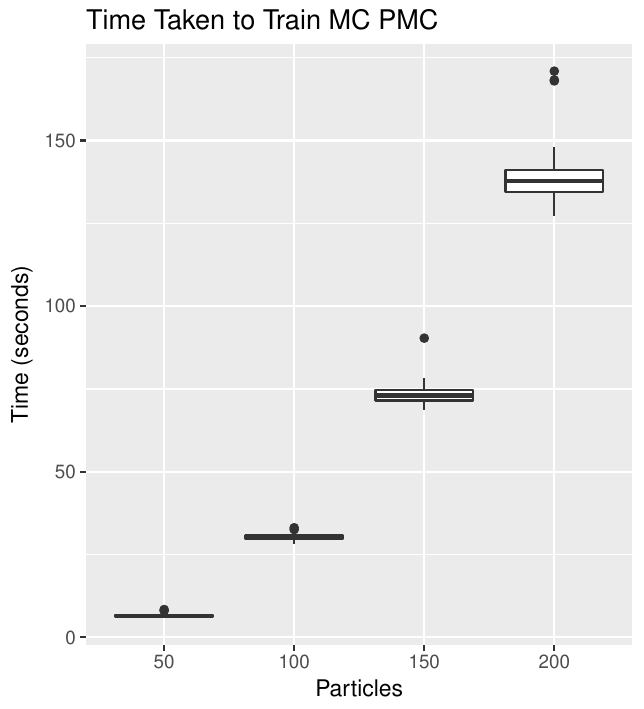}
\end{figure}

After conducting the K-L divergence analysis above, one might predict that Classification-PMC will perform better than LFIRE-PMC, since PMC is approximated solely through particle weights. We now directly compare performance within the PMC framework to show this: Figure 3 shows the mean squared error (MSE), averaged over the 5 $\mu_{i}$, for a total of 10 iterations across 100 datasets of single observations from a 5-dimensional multivariate Gaussian distribution with mean vector $(1,2,3,4,5)$ and an identity covariance matrix. The MSE refers to the mean squared error between the weighted average of the particles at iteration $i$ and the observation $Y$. To construct the classifiers, the likelihood function was simulated from 1000 times for each parameter.

\begin{figure}[p]
    \centering
    \caption{The mean squared error for the weighted means of the particles at each iteration of PMC, for both LFIRE and Classification-PMC. We conclude that Classification-PMC has a faster convergence rate than LFIRE.}
    \includegraphics[trim=0cm 0cm 1.7cm 0cm,width=1\linewidth]{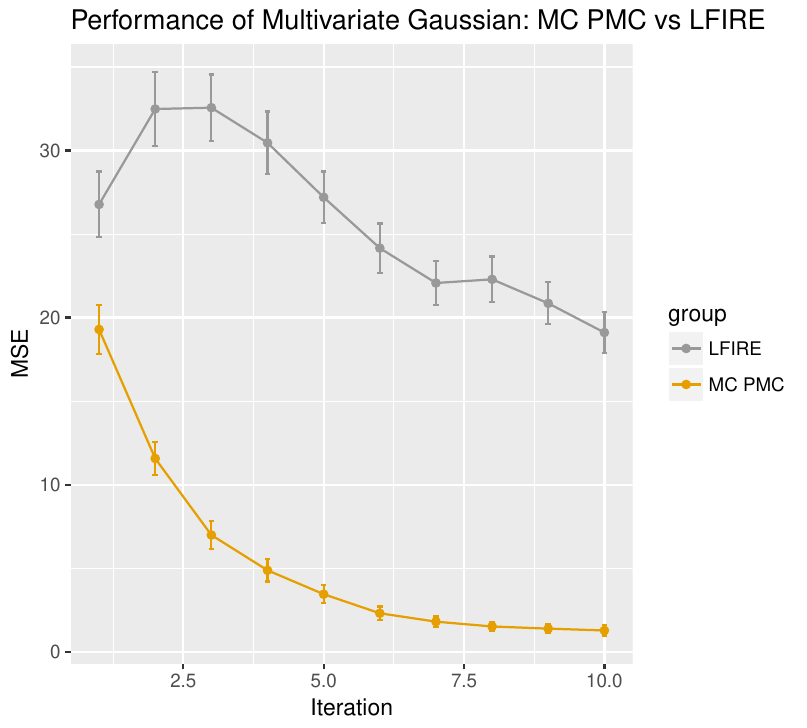}
\end{figure}

As the K-L divergence analysis indicated, there is faster convergence with Classification-PMC than LFIRE-PMC. The quality of the approximation to the weights is seen as early as the first iteration, and in the second iteration the importance of the approximation quality drastically affects convergence; the MSE for the multi-class classification approach decreases sharply just as the MSE for LFIRE increases sharply.

\subsection{Comparison to SMC ABC}

\subsubsection{Classification Methods}

\textbf{Artificial Neural Network:} A two-layer neural network was used to train the multi-class classifier, with 5 input units, 16 hidden units per hidden layer, and $N$ output units corresponding to the model class probabilities. The ReLU activation function was used for all layers except the last, which used the softmax function, a generalisation of the sigmoid function to multiple classes. The model was trained using the Adam optimiser, and with dropout and early stopping to prevent overfitting. A neural network was chosen because it can perform powerful classification, and hence highlights the benefits of Classification-PMC over SMC ABC. A full specification and explanation of the classifier is beyond the scope of this work, but illustrative code may be found in the Appendix (7.2.1).

\subsubsection{Results}

\begin{figure}
    \centering
    \caption{The root mean squared error between the weighted mean of particles and the true parameter values obtained using SMC ABC (grey) and Classification-PMC (orange). Classification-PMC uses fewer total simulations than SMC ABC and obtains similar accuracy, which is seen on the density plots for the total simulations and RMSE respectively.}
    \includegraphics[width=1\textwidth]{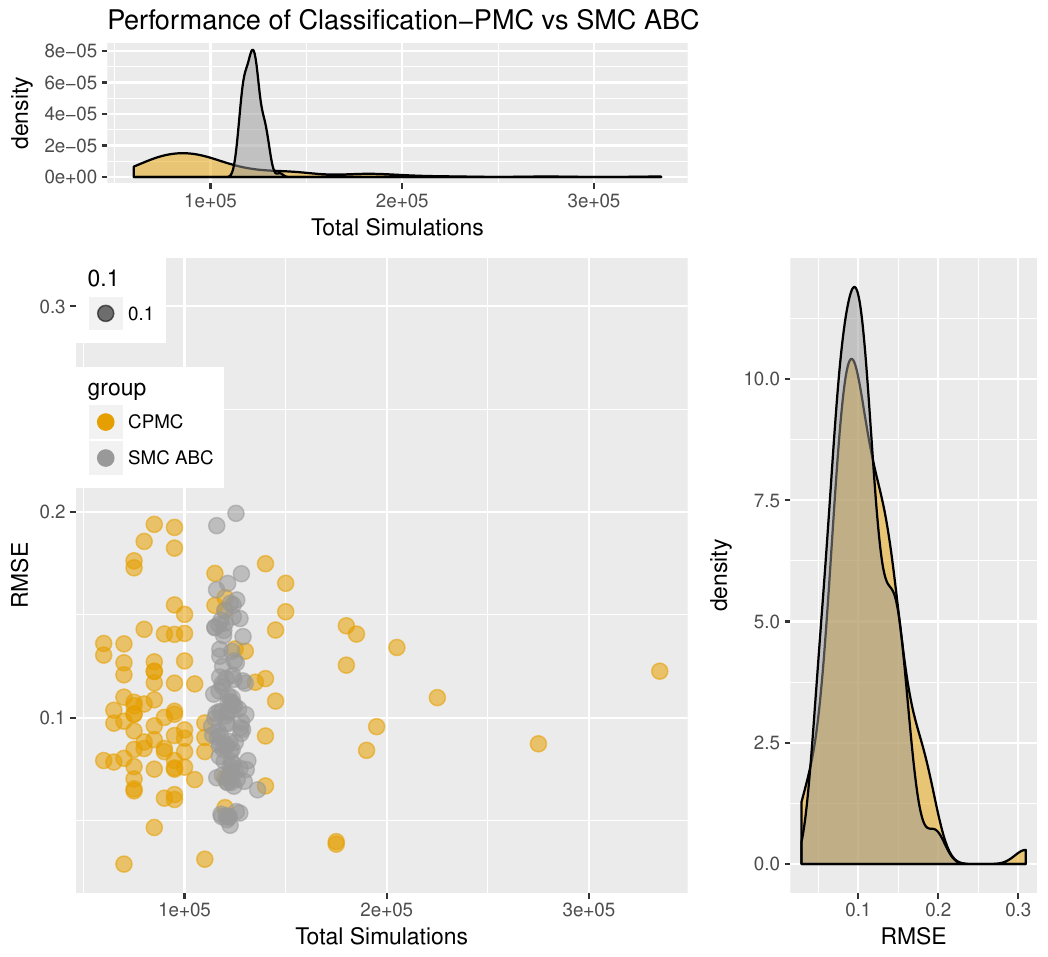}
\end{figure}

\begin{figure}
    \centering
    \caption{The root mean squared error between the weighted mean of particles and the true parameter values, obtained using SMC ABC (grey) and Classification-PMC (orange). Classification-PMC uses the same median total simulations as SMC ABC and obtains more accurate results, which is seen on the density plots for the total simulations and RMSE respectively.}
    \includegraphics[width=1\textwidth]{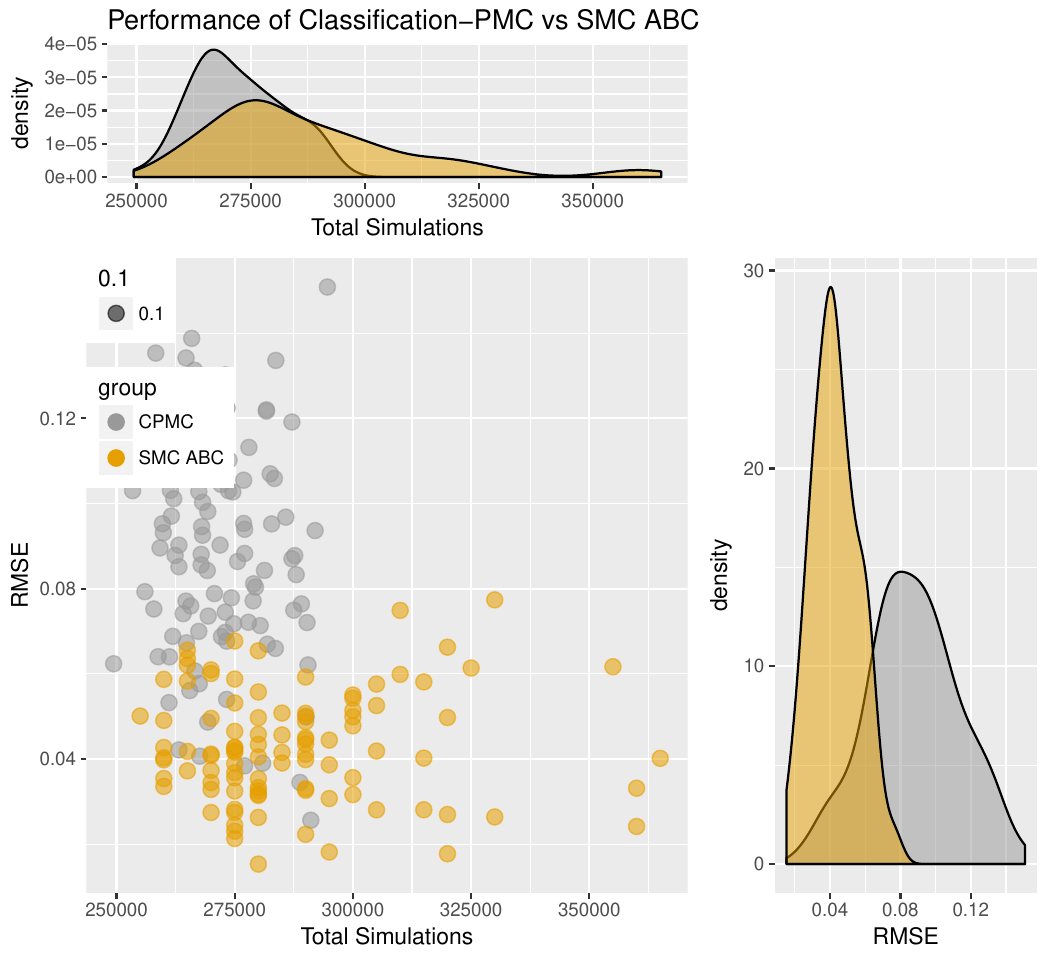}
\end{figure}

We now compare Classification-PMC to the state-of-the-art adaptive proposal approach: SMC ABC. For the implementation of Classification-PMC, 50 particles and 100 simulations per particle were used for the classification. Only particles post-convergence were used to calculate summary statistics, and we assumed convergence when the weighted mean and variance of the particles for each iteration held close to constant. The Engine for Likelihood-Free Inference (ELFI) Python package \cite{kangasraasio2016elfi} was used for the implementation of SMC ABC, which handles SMC ABC and other ABC methods. 10 iterations with 500 particles per iteration were used. 
Figure 4 and 5 compare the root mean squared error (RMSE) of the two methods on the same 100 datasets of observations, for two similar realisations of the total number of simulations from the likelihood function. The displayed results show SMC ABC using the best tolerance-schedule after 10 iterative improvements.

In Figure 4, we see that Classification-PMC uses fewer simulations than SMC ABC on average, with a similar RMSE density plot. For the second comparison in Figure 5, Classification-PMC has the same median total simulations as SMC ABC, albeit with a heavier tail, and the RMSE is far superior. This is strong evidence that Classification-PMC is more efficient than SMC ABC in the number of simulations from the likelihood. However, there is a computational cost in training the classifier for Classification-PMC, which is not taken into account by considering only simulations from the likelihood. For models where the likelihood is computationally difficult to simulate from, the above analysis is accurate, since the bulk of the computation is in simulating from the likelihood. It is for such models that Classification-PMC offers the greatest benefit.

\section{Conclusion}

The core contribution of this work is the proposition of a novel method, Classification-PMC, which blends adaptive proposals and classification to efficiently produce samples from the posterior distribution without the subjectivity that is encountered in most ABC methods. Classification-PMC uses classification and the resulting model-class probabilities to approximate particle weights in a population Monte Carlo scheme, enjoys asymptotic unbiasedness of samples, and uses all samples from all iterations after convergence has been reached. 

In our experiments, Classification-PMC produces more accurate particle-weight estimates than the state-of-the-art classification approach, LFIRE, while retaining its automatic selection of summary statistics. Classification-PMC was shown to be more efficient than the state-of-the-art adaptive proposal approach: SMC ABC. However, Classification-PMC incurs an additional computational cost in fitting the classification model, so the method will have better relative performance when it is difficult to simulate from the likelihood, or if the classification is made suitably cheap. Given more time, an interesting research direction is the exploration of more methods that sit between simulation and classification, which includes modifying the classifier used for Classification-PMC to trade off classification accuracy and total time taken to produce samples. This would require a more in depth analysis of total time taken for each method, as opposed to simply recording the total number of simulations from the likelihood.

\bibliographystyle{plainnat}
\bibliography{abc_phd.bib}

\end{document}